\makeatletter
\def\cl@chapter{}
\makeatother
\RequirePackage{fix-cm}
\documentclass[smallextended,envcountsect]{svjour3}       % onecolumn (second format)
\smartqed  % flush right qed marks, e.g. at end of proof

\usepackage[hidelinks]{hyperref}
\usepackage{amssymb}
\usepackage{graphicx}
\usepackage{amsfonts}       % blackboard math symbols
\usepackage{stmaryrd}		% Necessary for \ogreaterthan symbol
\usepackage{nicefrac}       % compact symbols for 1/2, etc.
\usepackage{microtype}      % microtypography
\usepackage{subcaption}
\usepackage[fleqn]{amsmath}

\usepackage{xspace}
\usepackage{siunitx}
\usepackage{float}
\usepackage{xcolor}
\usepackage{algorithm}
\usepackage{subfloat}
\usepackage[numbers,sort&compress]{natbib}
\newcolumntype{L}[1]{>{\raggedright\let\newline\\\arraybackslash\hspace{0pt}}p{#1}}
\newcolumntype{C}[1]{>{\centering\let\newline\\\arraybackslash\hspace{0pt}}p{#1}}
\newcolumntype{R}[1]{>{\raggedleft\let\newline\\\arraybackslash\hspace{0pt}}p{#1}}
\usepackage{enumitem}
\usepackage{booktabs}
\usepackage{listings}
\usepackage{color}

\usepackage{cleveref}
\usepackage{listings}
\usepackage[noend]{algpseudocode}

\usepackage{tikz}
\usepackage{pgfplots}

\pgfplotsset{compat=newest}
\usepackage{pgfplotstable}
\pgfplotsset{%
	legend style={font=\small},
	label style={font=\small},
	tick label style={font=\scriptsize},
	no markers,
	width=\linewidth,
	legend cell align=left,
	every axis/.append style={line width=0.75pt},
}
\usetikzlibrary{matrix}

\DeclareRobustCommand\sbseries{\fontseries{b}\selectfont\fontsize{8.5}{9.5}}

\newrobustcmd{\B}{\sbseries }

\begin{document}

\title{Integrating Human Knowledge Through Action Masking in Reinforcement Learning for Operations Research}

\titlerunning{Integrating Human Knowledge Through Action Masking in RL}        % if too long for running head

\author{Mirko Stappert \and Bernhard Lutz \and Niklas Goby \and Dirk Neumann}

\authorrunning{Stappert et al.} % if too long for running head

\institute{M.~Stappert, B.~Lutz, N.~Goby, D.~Neumann \at University of Freiburg, Rempartstr.~16, 79098 Freiburg, Germany\\
            \email{mirko.stappert@is.uni-freiburg.de, bernhard.lutz@is.uni-freiburg.de, niklas.goby@googlemail.com,  dirk.neumann@is.uni-freiburg.de}
            }

\date{April 3, 2025} %/ Accepted: }
% The correct dates will be entered by the editor

\maketitle

\begin{abstract}
Reinforcement learning (RL) provides a powerful method to address problems in operations research. However, its real-world application often fails due to a lack of user acceptance and trust. A possible remedy is to provide managers with the possibility of altering the RL policy by incorporating human expert knowledge. In this study, we analyze the benefits and caveats of including human knowledge via action masking. While action masking has so far been used to exclude invalid actions, its ability to integrate human expertise remains underexplored. Human knowledge is often encapsulated in heuristics, which suggest reasonable, near-optimal actions in certain situations. Enforcing such actions should hence increase trust among the human workforce to rely on the model's decisions. Yet, a strict enforcement of heuristic actions may also restrict the policy from exploring superior actions, thereby leading to overall lower performance. 
We analyze the effects of action masking based on three problems with different characteristics, namely, paint shop scheduling, peak load management, and inventory management. Our findings demonstrate that incorporating human knowledge through action masking can achieve substantial improvements over policies trained without action masking. In addition, we find that action masking is crucial for learning effective policies in constrained action spaces, where certain actions can only be performed a limited number of times. Finally, we highlight the potential for suboptimal outcomes when action masks are overly restrictive.

% Businesses can leverage these insights in similar problems to implement RL solutions that align with operational constraints and organizational expertise, enabling a more robust integration of RL in real-world settings.

% Our study extends the application of action masking by showing how it can be used to incorporate human knowledge into RL. This has significant implications for managers and decision-makers, as it provides a practical approach to improving RL policy effectiveness while addressing safety and trust concerns. 


\end{abstract}

\raggedbottom

\section{Introduction}

%% was ist RL + das funktioniert gut

Reinforcement Learning (RL), a branch of machine learning, has been successfully applied to solve complex problems such as the games of Chess and Go by self-play only \citep{Silver.2017,Silver.2018}. The resulting performance even surpasses the skills of human experts by significant amounts. In RL, an agent interacts with an environment to maximize the expected reward \citep{Sutton.2018}. The environment presents a numerical representation of the current state to the agent, who then decides upon the action to perform next. After receiving an action, the environment transforms to the next state and emits a numerical reward to the agent. Advances in deep reinforcement learning (DRL) allowed researchers to solve the problem of the exploding state space as the mapping of states to Q-values or action probabilities is now approximated by neural networks \citep{Mnih.2015}.

%Therefore, businesses could pretrain an RL policy in a simulated environment with known problem parameters and subsequently apply it to similar but unseen problems in real-world applications.

%% advantage

%put in Intro: \subsection{Reinforcement Learning in Operations Research}

%% Überblick zu den Problemtypen: ABCD mit Review Papieren

Over the past few years, RL has also been increasingly used as a powerful tool for addressing complex problems in operations research (OR). The considered problem areas include production planning and control \citep{Esteso.2023,Panzer.2022,Zhang.2024}, supply chain management and inventory management \citep{Rolf.2023}, machine scheduling \citep{Kayhan.2023}, machine maintenance planning \citep{Ogunfowora.2023}, and quality control \citep{Paraschos.2024}, among many others. 
A major advantage of RL over traditional solution methods (e.g., hand-crafted rules, heuristics, mathematical solvers) is that a trained RL policy presents a fast and adaptive solution method. Businesses often encounter similar decision problems where the parameters follow known distributions like package delivery problems in the same district \citep[e.g.,][]{Bengio.2021,Schiewe.2024}, or sequencing problems with similar demand plans \citep[e.g.,][]{Brammer.2022a,Brammer.2022b}. A RL policy can be trained in a simulated environment and subsequently transfer knowledge from prior learning experiences to unseen problem instances \citep{Taylor.2009}. By contrast, the aforementioned traditional solution methods are designed to solve each problem instance individually without relying on prior experience. In addition, RL can be trained to account for the uncertainty in real-world operations regarding delivery times or customer demand. The paradigm of RL hence aligns particularly well with the goals of Industry 4.0, where adaptive and fast decision-making under uncertainty are crucial \citep{Choi.2022,Weckenborg.2024}.

%% anfangs nur als Tool um heuristic auszuwählen... hier 3-4 Paper auflisten --> nur max so gut wie die heuristic

%% RL direkt als Lösungsmechanismus --> übersetze ILP zu MDP --> schwierig da man X Sachen beachten muss wie
%% valid solutions + anderes: stochastic multi-objective ... sonstige constraints
%% --> negative rewards als punishment oder action masking

A major challenge of implementing RL in real-world operations is the task of converting the problem presumably formulated as a mixed-integer linear programming (MILP) problem to a Markov decision process, including states, actions, and reward function, that is suitable for RL policy learning \citep{Sutton.2018}. Here, businesses need to ensure that i) the policy generates valid solutions, and that ii) a (near-)optimal policy of the MDP also generates (near-)optimal solutions of the actual MILP problem. A common approach to guiding the RL policy towards generating only valid solutions is given by returning the same state and penalizing invalid actions with large negative rewards \citep[e.g.,][]{Brammer.2022b,Valet.2022}. While this method generally achieves the desired outcome, the policy will still perform invalid actions during the training process, thereby increasing the time required for learning effective policies. An alternative is given by action masking \citep[e.g.,][]{Huang.2020,Luo.2022,Wang.2024}, a technique in RL that constrains the action space by limiting the set of available actions in specific states.

%% algorithm aversion --> human knowledge

In addition to these technical challenges, the implementation of RL, like other AI-based solution methods, often fails in real-world operations due to a lack of human trust and acceptance of AI decisions \citep{Dietvorst.2015,Lehmann.2022}. Although problems from OR are often considered as repetitive low-stake problems \citep{Notz.2024}, human decision-makers still want to be able to make appropriate adjustments, particularly when models are wrong \citep{Dietvorst.2018}. Therefore, a possible remedy to increase trust and user acceptance is given by allowing managers to modify the RL policy by including human domain knowledge. Human knowledge can prescribe reasonable heuristic actions in certain states, or even provably optimal actions. By ensuring that a policy incorporates human knowledge, the workforce should be less likely to suffer from algorithm aversion and instead rely on the model's decisions.

%% what-we-do

In this study, we analyze the inclusion of human domain knowledge via action masking. While traditionally used to prevent invalid actions, its potential for enforcing heuristic-suggested or provably optimal actions remains unexplored. Although a RL policy could, in theory, learn the actions suggested by the action mask implicitly after sufficient training time, there is no such guarantee in DRL due to the ``deadly triad'' of function approximation, bootstrapping, and off-policy learning \citep{Hasselt.2018,Sutton.2018}. Unlike tabular-based Q-learning, where convergence to a global maximum is theoretically guaranteed \citep{Sutton.2018}, DRL approximates the values of states using a neural network, introducing function approximation errors. In fact, an update of the network's weights influences several states instead of only a single state-action entry of the Q-value table. In addition, policy updates depend on previous estimates (bootstrapping) and are often applied to past experience (off-policy learning), thereby possibly amplifying approximation errors. Including action masks in the training process could hence lead to considerably superior policies. At the same time, action masks might also restrict the flexibility of the policy, preventing it from reaching particular optima.

%% empirical

We examine the benefits and caveats of action masking based on three OR problems with different characteristics (paint shop scheduling, peak load management, inventory management). In particular, we outline the effects of action masking when incorporating different types of human knowledge regarding heuristic and provably optimal actions. Our evaluation yields several important insights. First, we find that action masking can lead to considerable increases in the performance of a trained RL policy. Enforcing heuristic actions in a paint shop scheduling problem yields considerably better policies than omitting action masking. Second, we observe that, for problems with constrained action spaces, applying action masking can even be necessary to learn effective policies. The considered peak load management only allows a limited number of turn-off operations. If the policy is not guided using human knowledge, it never learned to effectively manage the load. Third, we acknowledge that action masking should be applied with caution as the enforcement of non-optimal heuristics may also harm the resulting performance.

% Contribution

Our study contributes to the literature by analyzing the effects of incorporating human domain knowledge into the training and application of RL policies via action masking. Alternatives to action masking include imitation learning \citep{Hussein.2017} and inverse RL \citep{Arora.2021}, allowing the policy to learn from observing human actions. However, it may not often be possible to learn from human actions due to labor shortage or the problem is simply too complex for a human to solve optimally. Other approaches employ reward shaping \citep{Cheng.2021} or transfer learning \citep{Bianchi.2015}. Yet, these methods still do not allow human decision-makers to enforce or disallow actions in certain situations. While our study is limited to offline learning, where a policy is pretrained in a simulated environment, action masking may also help to counter the cold start problem that occurs when RL is trained directly in real-world applications. Here, action masking can be used to prevent the execution of clearly unreasonable actions, as, e.g., reordering stock when the current inventory is full, which would result in considerable losses. We thus hope to pave the way for future studies that develop RL-based solution methods for OR problems.

%In addition, our findings based on three case studies of different OM problems suggest useful guidelines for decision-makers opting to implement RL in real-world operations. 

%% structure

The remainder of this paper is structured as follows. \Cref{sec:rlam} describes the methodological framework by introducing definitions of Markov decision processes and detailing how human domain knowledge can be included via action masking. 
%\Cref{sec:rw} provides an overview of related work regarding RL in operations in general and how action masking was used in prior studies. 
\Cref{sec:paintshop} presents the paint shop scheduling problem, utilizing action masking to implement multiple heuristics.
\Cref{sec:loadmanagement} explores the peak load management problem with constrained actions. 
\Cref{sec:inventorymanagement} examines an inventory management problem with stochastic demands and delivery times.
\Cref{sec:discussion} discusses our findings and provides an outlook on future research.

%% Framework

\section{Methodological Framework}
\label{sec:rlam}

\subsection{Reinforcement Learning}

Reinforcement learning is a type of machine learning where an agent interacts with an environment to learn a policy, which specifies the optimal action(s) in a given state. Most real-world decision problems with known parameters and a finite number of actions can be transferred to a problem from RL. RL problems are formally modeled as a Markov decision process \citep{Sutton.2018}.
\begin{definition}[Markov decision process]
\label{def:mdp}
A Markov decision process (MDP) is a tuple $(S, A, T, R, s_0, \gamma)$ with state space $S$, action space $A$, transition function $T(s_{t+1} \mid a, s_t)\in [0,1]$, reward function $R(r_t \mid a, s_t)\in  [0,1]$, initial state $s_0$ and discount factor $\gamma \in [0,1]$.
%We consider a finite time horizon with a maximum of $H$ timesteps. $A(s_t) \subseteq A$ denotes the set of available actions in state $s_t \in S$.
\end{definition}

%% ein Absatz wie das funktioniert

The state $s_t \in S$ reflects the current situation of the environment. The action space $A$ denotes all possible decision options (e.g., ordering supplies, producing a certain item). The transition function $T(s_{t+1} \mid a, s_t)$ specifies how the environment evolves from one state to another if a particular action is performed. The transition function can be deterministic (e.g., in production scheduling problems with fixed processing times) or stochastic (e.g., in inventory management problems with demand uncertainty). In the general stochastic case, $T(s_{t+1} \mid a, s_t)$ denotes the probability that $s_{t+1}$ follows $s_t$ given that action $a$ is performed. The reward function can also be stochastic, so that $R(r_t \mid a,s_t)$ specifies the probability of receiving reward $r_t$ when action $a$ is performed in state $s_t$. The reward function should be designed in a way that assigns positive rewards to actions that are effective in achieving the overarching goal (e.g., fulfilling demand while minimizing storage costs) and negative rewards to ineffective and sub-optimal actions.

%% policy definieren

The goal of RL is to find a policy $\pi_\theta(a|s_t)$, which maximizes the expected total sum of discounted rewards $J=\mathbb{E}_{\pi_\theta}\left[\sum_{t=0}^H \gamma^t r_t \right]$. Specifically, $\pi_\theta(a|s_t)$ denotes the probability of performing action $a$ in state $s_t$. In deep reinforcement learning, a neural network receives the state as input and outputs a probability distribution over the action space. The policy parameters $\theta$ denote the weights of the neural network, which must be learned during the training process. The training process alternates between generating experience in the form of state-action-reward tuples $s_0,a_0,r_0,\dots,s_{T},a_{T},r_{T}$ by sampling from the policy and using this experience to update the policy parameters $\theta$. 

%% Überleitung actionmasking durch Problem mit statischem action space erklären

So far, it is assumed that the action space is constant in every state, i.e., each action can always be performed. However, in real-world applications, only a subset of all actions are admissible in each specific state. Invalid actions are often excluded by introducing $A(s_t)$ as the set of admissible actions in $s_t$. While introducing $A(s_t)$ is theoretically feasible, there are practical limitations. In fact, the policy networks employed in deep RL have a fixed number of output neurons, which cannot be adjusted dynamically during the training process. Consequently, invalid actions must be handled manually during the training process and real-world application. A common approach is to mask invalid actions by setting their probability to zero and sampling from the remaining (admissible) actions. 

%% action masking training process

While action masking is an obvious step in the application of a trained policy, it also affects the training process in general. First, action masking prevents the exploration of invalid actions, which increases the efficiency of the training process. Second, action masking influences learning by updating the policy parameters according to an adjusted policy gradient \citep{Huang.2020}. Third, action masking provides an early guideline for an untrained RL policy so that it suffers less from the cold start problem, where obviously non-optimal actions are explored in the early learning episodes. From an implementation view, action masking has recently been implemented as part of the popular RL framework ``Stable Baselines3'' \citep{raffin2021stable} based on the implementation of \cite{Huang.2020}.

\subsection{Integrating Human Domain Knowledge via Action Masking}

%% Definition and dynamic action spaces

An action mask $m$ is a function $m{:}\ S\times A\rightarrow\{0,1\}$ that reduces the set of actions in a given state \citep{Huang.2020}.  By evaluating $m(a, s_t)$ for all actions, we obtain 
\begin{equation}
A(s_t)=\{a\in A\mid m(a, s_t)=1\}.
\end{equation}
Action masking can hence be used to reduce the set of admissible actions in reinforcement learning. In addition to excluding invalid actions, the functionality of action masking can be employed to include human domain knowledge, namely, (i)~prescribe heuristic rules and, more strictly, (ii)~enforce provably optimal actions.

%% invalid actions

To exclude invalid actions, we simply set $m(a, s_t)=0$ if $a$ is invalid in $s_t$ \citep{vinyals2017starcraft, berner2019dota, Huang.2020} 
\begin{equation}
\label{eq:invalid}
m(a, s_t)=\begin{cases}
1 & \text{if } a \text{ is valid in } s_t\\
0 & \text{else}.
\end{cases}
\end{equation}

To the best of our knowledge, action masking has so far only been used to exclude invalid actions \citep[e.g.,][]{vinyals2017starcraft, berner2019dota, Huang.2020}. In fact, the term ``action masking'' is often used in combination with ``invalid'' like in the title ``A Closer Look at Invalid Action Masking in Policy Gradient Algorithms'' by \citet{Huang.2020}. 
Nevertheless, the application of action masking is equally suited to guiding the policy towards heuristic actions as to enforcing valid actions.

\subsubsection*{Prescribing Domain Heuristics}

For many problems from operations research, there are certain heuristics that can be used to restrict the action space. For instance, in an inventory management problem, there might be knowledge that we should produce at most twice the demand of the previous period. While producing more is still possible, we know that many of the products will not be sold in the next period. Therefore, the heuristic suggests that the action space is restricted to actions that order less than twice the demand of the previous period. 

%% heuristic formal

Let $h(s_t)$ denote the action suggested by the given heuristic in state $s_t$. We do not strictly enforce a heuristic, as this would make the policy equal to the heuristic. Instead, we aim to approximately prescribe the heuristic, thereby imparting knowledge to the RL agent but also giving it the flexibility to deviate from the action suggested by the heuristic.

%% structure

Prescribing a heuristic while providing the RL policy with a limited amount of flexibility is, in particular, possible if the action space has a certain structure that allows us to rank actions based on a reasonable order. For instance, in inventory management problems, this corresponds to the quantity of items to be produced in the next period. Given such an order of actions, we can restrict the action space to the action suggested by the heuristic and the actions close to $h(s_t)$ according to a given threshold $M$
\begin{equation}
\label{eq:prescribe_approx_distance}
m(a, s_t)=\begin{cases}
1 & \text{if } \mid h(s_t)-a \mid \leq M \\
0 & \text{else}.
\end{cases}
\end{equation}

Similarly, the action space can be restricted to actions greater or smaller than the action suggested by the heuristic
\begin{equation}
\label{eq:prescribe_approx_ordered}
m(a, s_t)=\begin{cases}
1 & \text{if } a\geq h(s_t) \text{ (conversely } a \leq h(s_t) \text{)}\\
0 & \text{else}.
\end{cases}
\end{equation}
Note that the specific selection of the action mask depends on the particular problem characteristics.
%Which of those design choices should be taken of course depends on the specific problem characteristics.

\subsubsection*{Enforcing Optimal Actions}
\label{sec:enforce_optimal}

The strictest approach of incorporating human domain knowledge is given by enforcing provably optimal actions. An action $a^*$ is optimal in state $s_t$ if there is an action sequence that maximizes cumulative reward in $s_t$ and starts with $a^*$. Due to their optimality, these actions should be enforced while disallowing all non-optimal actions. However, we generally know the optimal actions only for a small subset of states $S' \subset S$. 
%\footnote{We can also proceed similarly in the case that we do not know which actions are optimal, but are only aware of which options are non-optimal.}
We thus set $m(a^*, s_t)=1$ for all optimal actions $a^*$ and $m(a, s_t)=0$ for all non-optimal actions in all states $s_t \in S'$. For all other states $s_t\in S\setminus S'$, where optimal actions are not known, we simply allow all actions. Taken together, we get
\begin{equation}
\label{eq:prescribe_optimal}
m(a, s_t)=\begin{cases}
1 & \text{if } s_t\in S' \text{ and } a \text{ is optimal in } s_t\\
0 & \text{if } s_t\in S' \text{ and } a \text{ is not optimal in } s_t\\
1 & \text{if } s_t\in S\setminus S'.
\end{cases}
\end{equation}

\subsubsection*{Combining Action Masks}

%% Combining action masks

So far, we only discussed individual action masks.
However, there might be situations where several reasonable action masks should be combined. 
Therefore, we outline two approaches to combine action masks. First, we define the conjunction operation $m_1 \oplus m_2$, which only allows action $a$ to be performed in state $s_t$, if both masks allow $a$
\begin{equation}
\label{eq:combine_add}
(m_1 \oplus m_2)(a, s_t)=\begin{cases}
			1 & \text{if } m_1(a, s_t)=1 \text{ and } m_2(a, s_t)=1 \\
			0 & \text{else}.	
			\end{cases}
\end{equation}

The second operation $m_1 \ogreaterthan m_2$ applies two action masks in a sequential way, while assigning higher priority to $m_1$. This approach is particularly relevant if we want to prioritize one mask over another. $m_1 \ogreaterthan m_2$ thus applies $m_1$ if $m_1$ is \emph{active} in $s_t$, i.e., $m_1$ forbids at least one action, and $m_2$ otherwise
\begin{equation}
\label{eq:combine_seq}
(m_1 \ogreaterthan m_2)(a, s_t)=\begin{cases}
			m_1(a, s_t) & \text{if } \exists a' \in A{:} \ m_1(a', s_t) = 0 \\
			m_2(a, s_t) & \text{else}.			
			\end{cases} 
\end{equation}

\subsubsection*{Policy Gradient Calculations with Action Masking}

The implementation of action masking by \citet{Huang.2020} builds upon the policy gradient method. The objective of maximizing the expected sum of discounted rewards $J=\mathbb{E}_{\pi_\theta}\left[\sum_{t=0}^H \gamma^t r_t \right]$ is achieved by applying gradient ascent optimization to the function $J$ with respect to the policy parameters $\theta$. The calculation of the policy gradient is enabled by the policy gradient theorem \citep{Sutton.2018}
\begin{equation}
\nabla_\theta J = \mathbb{E}_{\pi_{\theta}}\left[\sum_{t=0}^H \nabla_\theta \log \pi_\theta(a| s_t) G_t \right],
\end{equation}
where $G_t = \sum_{k=0}^H \gamma^k r_{t+k}$ is the sum of discounted rewards starting at time step $t$. The gradient $\nabla_\theta J$ can be estimated by sampling trajectories from the policy $\pi_\theta$ and averaging the resulting values inside the expectation operator. 
The action probabilities are calculated via the softmax function from the logits $l_\theta(a|s_t)$ as
\begin{equation}
\pi_\theta(a|s_t) = [\text{softmax}(l_\theta(\cdot|s_t))]_a= \frac{\exp(l_\theta(a|s_t))}{\sum_{a'\in A}\exp(l_\theta(a'|s_t))}
\end{equation}

%% mit action mask

Given an action mask $m$, one proceeds as follows. First, the logits are updated by setting actions with $m(a, s_t)=0$ to negative infinity
\begin{equation}
l^m_\theta(a|s_t) = \begin{cases}
				l_\theta(a|s_t) & \text{if } m(a, s_t)=1 \\
				-\infty  & \text{if } m(a, s_t)=0.
			\end{cases} 
\end{equation}
Second, the action probabilities are recalculated based on the adjusted logits 
\begin{equation}
\pi_\theta^m(a|s_t) = [\text{softmax}(l^m_\theta(\cdot|s_t))]_a.
\end{equation}
The effect of action masking can easily be shown on a small example. Let the logits of state $s_t$ be given as $[l_\theta(a_0|s_t),l_\theta(a_1|s_t),l_\theta(a_2|s_t)]=[1,1,1]$. The three actions $a_0,a_1,a_2$ are thus performed with equal probability of $\frac{1}{3}$. Given an action mask $m$ that disallows action $a_2$ in $s_t$, the resulting probabilities based on the adjusted logits are $\pi^m_\theta(a_0|s_t)=\pi^m_\theta(a_1|s_t)=\frac{1}{2}$ and $\pi^m_\theta(a_2|s_t)=0$. 

Finally, the modified policy gradient $\nabla^m_\theta J$ is calculated using the updated action probabilities of $\pi^m_\theta(a|s_t)$
\begin{equation}
\nabla^m_\theta J = \mathbb{E}_{\pi_{\theta}}\left[\,\sum_{t=0}^H \nabla_\theta \log \pi^m_\theta(a| s_t) G_t \right].
\end{equation}

An action mask is thus not only enforced by hand but also incorporated into the learning process itself by performing gradient ascent steps with an adjusted policy gradient.

\section{Problem 1: Paint Shop Scheduling}
\label{sec:paintshop}

The paint shop scheduling problem deals with minimizing the number of color changes in the painting procedure of the automotive manufacturing process. Therefore, an incoming sequence of unordered cars with assigned colors has to be reshuffled using a multi-lane buffer system that allows storage and retrieval operations.

%%%%%%%%%%%%%%%%%%%%%%%%%%%%%%
%%%% COLOR DEFINITIONS %%%%%%%
%%%%%%%%%%%%%%%%%%%%%%%%%%%%%%

\definecolor{blue}{rgb}{0.36, 0.54, 0.66}
\definecolor{red}{rgb}{0.8, 0.0, 0.0}
\definecolor{green}{rgb}{0.0, 0.5, 0.0}
\definecolor{lila}{rgb}{0.74, 0.2, 0.64}

\begin{figure}[H]
\centering
\begin{tikzpicture}[scale=0.75]
\tikzstyle{every node}=[font=\scriptsize,align=center,execute at begin node=\setlength{\baselineskip}{11pt}]
% current color

    \node[text width=3cm,align=center] at (7.6,4.2) {color of outgoing sequence};
    \draw[->,thick] (7.6,3.7) -- (7.6,2.6);
    
  % Draw line of squares with e^t_n
    \node[] at (-2.5,2) {...};
    %\draw[fill=blue,fill opacity=0.6,
    %text opacity=1]  (0,1.5) rectangle ++(1,1) node[midway][scale=1] {$e_{t,1}$};
    \draw[fill=green,fill opacity=0.6,
    text opacity=1] (-1,1.5) rectangle ++(1,1) node[midway][scale=1] {$e_{t,1}$};
    \draw[fill=red,fill opacity=0.6,text opacity=1] (-2,1.5) rectangle ++(1,1) node[midway][scale=1] {$e_{t,2}$};

    % Brace with description for incoming sequence
    \draw [decorate,
    decoration = {brace, mirror, amplitude=5pt}] (-3,1.3) --  (0,1.3);
    \node[] at (-1.5,0.8) {Incoming sequence};
 
  % Draw big square with B^t_{i,j}
  \foreach \x in {0,...,4}
    \foreach \y in {0,...,3}
      \pgfmathtruncatemacro\i{4-\y}
      \pgfmathtruncatemacro\j{1+\x}
      \draw (1+\x,\y) rectangle ++(1,1) 
      node[midway][scale=1] {$B_{t,\i,\j}$};
      
   % first lane
	\draw[fill=blue,fill opacity=0.6,
    text opacity=1] (5,3) rectangle ++(1,1)node[midway][scale=1] {$B_{t,1,5}$};   
    \draw[fill=blue,fill opacity=0.6,
    text opacity=1] (4,3) rectangle ++(1,1)
    node[midway][scale=1] {$B_{t,1,4}$};  
	\draw[fill=red,fill opacity=0.6,
    text opacity=1] (3,3) rectangle ++(1,1)
    node[midway][scale=1] {$B_{t,1,3}$};
	\draw[fill=red,fill opacity=0.6,
    text opacity=1] (2,3) rectangle ++(1,1)
    node[midway][scale=1] {$B_{t,1,2}$};
    
    % second lane	  
	\draw[fill=blue,fill opacity=0.6,
    text opacity=1] (5,2) rectangle ++(1,1)node[midway][scale=1] {$B_{t,2,5}$};  
	\draw[fill=green,fill opacity=0.6,
    text opacity=1] (4,2) rectangle ++(1,1)
    node[midway][scale=1] {$B_{t,2,4}$}; 	
	
	% third lane	
	\draw[fill=lila,fill opacity=0.6,
    text opacity=1] (5,1) rectangle ++(1,1)
    node[midway][scale=1] {$B_{t,3,5}$}; 
	\draw[fill=lila,fill opacity=0.6,
    text opacity=1] (4,1) rectangle ++(1,1)
    node[midway][scale=1] {$B_{t,3,4}$}; 	
	\draw[fill=blue,fill opacity=0.6,
    text opacity=1] (3,1) rectangle ++(1,1)
    node[midway][scale=1] {$B_{t,3,3}$}; 
    
    % fourth lane	
	\draw[fill=blue,fill opacity=0.6,
    text opacity=1] (5,0) rectangle ++(1,1)
    node[midway][scale=1] {$B_{t,4,5}$}; 
	   
    % Brace with description for Buffer
    \draw [decorate,
    decoration = {brace, mirror, amplitude=5pt}] (1,-0.5) --  (5.9,-0.5);
    \node[] at (3.5,-1) {Buffer};
    
	% FIFO Queue   
	\draw[->,thick]  (1,4.5) -- (6,4.5);
    \node[text width=3cm, align=center] at (3.5,5) {FIFO queue};
    
    % Brace with description for Storage
    \draw [decorate,
    decoration = {brace, mirror, amplitude=5pt}] (0,-0.5) --  (0.9,-0.5);
    \node[] at (0.5,-1) {Storage};
    
    % Brace with description for Retrieval
    \draw [decorate,
    decoration = {brace, mirror, amplitude=5pt}] (6,-0.5) --  (6.9,-0.5);
    \node[] at (6.5,-1) {Retrieval};
    
  % Draw arrows for storage
  \foreach \x in {0,...,3}
  \pgfmathtruncatemacro\j{2-\x}
    \draw[->,thick] (0,2) -- (0.9,\x+0.5) node[pos = 0.6, above, sloped, inner sep=1pt] {};
      
  % Draw arrows for retrieval
  \foreach \x in {0,...,3}
   \pgfmathtruncatemacro\j{4-\x}
    \draw[->, thick] (6,\x+0.5) -- (7,1.8+0.1*\x) node[pos = 0.4, above, sloped, inner sep=1pt] {};
    \draw[->, ultra thick] (6,0.5) -- (7,1.8) node[pos = 0.4, above, sloped, inner sep=1pt] {};

    \node[text width=5cm] at (9,0) {$B_{t,4,5} = p_t \Rightarrow $ no color change};

   % Outgoing sequence
\node[] at (9.6,2) {...};
    \draw[fill=blue,fill opacity=0.6,
    text opacity=1] (7.1,1.5) rectangle ++(1,1) node[midway] {$p_t$};
    \draw[fill=lila,fill opacity=0.6,
    text opacity=1] (8.1,1.5) rectangle ++(1,1) node[midway] {};
    %\draw[fill=green,fill opacity=0.6, text opacity=1] (12.1,1.5) rectangle ++(1,1) node[midway] {};

   %\node[] at (13.5,2) {Outgoing sequence};
    \draw [decorate,
    decoration = {brace, mirror, amplitude=5pt}] (7.1,1.3) --  (10.1,1.3);
    \node[] at (8.6,0.8) {Outgoing sequence};
    
\end{tikzpicture}
\caption{Paint shop problem with a 4x5 buffer (four lanes of width five). The system retrieves from lane 4 without causing a color change.}
\label{fig:visualization_form}
\end{figure}
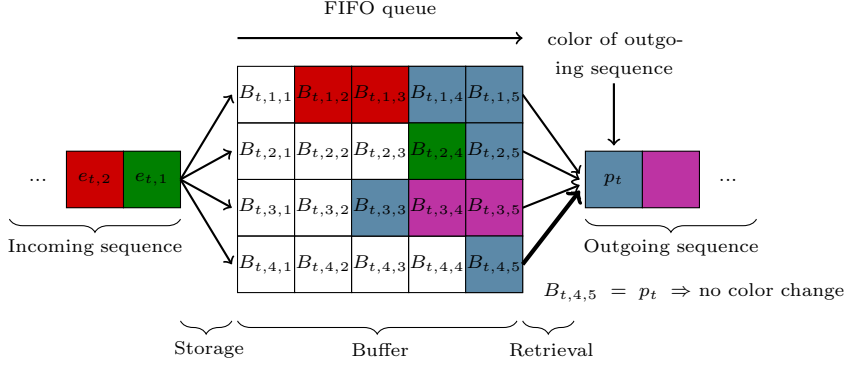

\subsection{Environment}
At each time step, the next car of the incoming sequence can be stored in the rightmost free position of a buffer lane or the rightmost car of a buffer lane can be retrieved and added to the last position of the outgoing sequence. The buffer consists of $L$ lanes, each of width $W$. 
The action space hence consists of $2L$ actions %, the first half corresponding to retrieve actions and the second half to storage.
\begin{equation} \label{state-space}
A^{\text{PS}} = \{\underbrace{1,\dots, L}_{\substack{\text{Retrieve}\\ \text{actions}}}, \underbrace{L+1,\dots,2L}_{\substack{\text{Store}\\ \text{actions}}}  \}.
\end{equation}
%To make these decisions the agent is provided with the current state, which contains the buffer content, the incoming sequence and the current painting color:
We define the state to contain the entire buffer content, the next $K=5$ colors of the incoming sequence, and the current color of the outgoing sequence
\begin{equation} \label{integer-state-representation}
s_t^{\text{PS}}= (\underbrace{B_{t,1,1},\dots,B_{t,L,W}}_{\text{Buffer content}}, \underbrace{e_{t,1},\dots,e_{t,K}}_{\substack{ \text{K next incoming cars} \\ \text{of input sequence}}}, \underbrace{p_t}_{\substack{ \text{Current} \\ \text{painting color}}}).
\end{equation}
Each value ($0,1,\dots, C$) is one-hot encoded and either denotes an empty position or a color $1,\dots,C$. 

%% reward function

The reward function penalizes invalid actions and retrieve actions causing a color change.
Invalid actions yield a reward of $-10$. Valid store actions and retrieve actions that cause a color change receive no reward.
However, valid retrieve actions not causing a color change are assigned a positive reward of $1$.
%\small
\begin{equation} \label{reward-function}
R^{\text{PS}}(a,s_{t})=\begin{cases}
 0 & a \leq L,\, B_{t,a,W}\neq 0,\, B_{t,a,W} \neq p_t \\
   & \text{(valid retrieval with color change)}\\
 1 & a \leq L,\, B_{t,a,W}\neq 0,\, B_{t,a,W}=p_t \\
   & \text{(valid retrieval without color change)}\\
 -10 & a \leq L,\, B_{t,a,W}=0 \quad \text{(invalid retrieval)}\\
 0 & a > L,\, B_{t,a-L,1}= 0,\, e_{t,1} \neq 0 \quad \text{(valid store)}\\
 -10 & a > L,\, B_{t,a-L,1} \neq 0 \text{ or } e_{t,1} = 0 \quad \text{(invalid store)}
\end{cases}
\end{equation}
%\normalsize
We implement the environment in the RL framework ``Stable Baselines3'' \citep{raffin2021stable}. The method for policy learning is proximal policy optimization \citep[PPO,][]{Schulman.2017}. The hyperparameters of PPO (e.g., size of policy network and PPO coefficients) are provided in Appendix A.

\subsection{Action Masking}

\begin{figure}[H]
\captionsetup[subfigure]{justification=centering,font=footnotesize,skip=0pt}
	\footnotesize
     \centering
     \begin{subfigure}[t]{0.47\textwidth}
         \begin{tikzpicture}[scale=0.65]
         
\tikzstyle{every node}=[font=\footnotesize,align=center,execute at begin node=\setlength{\baselineskip}{11pt}]
% next incoming color
    \node[align=center] at (0.5,3.3) {color of next \\ incoming car};
    \draw[->,thick] (0.5,2.9) -- (0.5,2.1);

% current color
    \node[] at (6.6,3.3) {current color};
    \draw[->,thick] (6.6,2.9) -- (6.6,2.1);
  % Incoming sequence
    \draw [fill=blue] (0,1) rectangle ++(1,1) node[midway] {};
     %Outgoing
\draw [fill=blue] (6.1,1) rectangle ++(1,1) node[midway] {};  
  % Big square with B^t_{i,j}
  \foreach \x in {0,...,2}
    \foreach \y in {0,...,2}
      \pgfmathtruncatemacro\i{3-\y}
      \pgfmathtruncatemacro\j{1+\x}
      \draw (2+\x,\y) rectangle ++(1,1) 
      node[midway] {};
% Color buffer	
	\draw[fill=blue] (3,2) rectangle ++(1,1);  
	\draw[fill=green] (4,2) rectangle ++(1,1);
%first lane
	\draw[fill=lila] (2,1) rectangle ++(1,1); 
	\draw[fill=lila] (3,1) rectangle ++(1,1); 	
	\draw[fill=blue] (4,1) rectangle ++(1,1);      
     % Arrows
   \foreach \x in {0,...,2}
    \draw[->,thick] (1,1.5) -- (2,\x+0.5);  
  \foreach \x in {0,...,2}
    \draw[->,thick] (5,\x+0.5) -- (6,1.4+0.1*\x); 
\end{tikzpicture}
\vspace{0.2cm}
         \caption{Invalid action mask: Only storing in lane 1 or 3 and retrieving from lane 1 or 2 is allowed.}
	     \end{subfigure}
	     \hspace{0.5cm}
     \begin{subfigure}[t]{0.47\textwidth}
%         \centering
         \begin{tikzpicture}[scale=0.65]
\tikzstyle{every node}=[font=\footnotesize, align=center,execute at begin node=\setlength{\baselineskip}{11pt}]
% next incoming color
    \node[align=center] at (0.5,3.3) {color of next \\ incoming car};
    \draw[->,thick] (0.5,2.9) -- (0.5,2.1);
% current color
    \node[] at (6.6,3.3) {current color};
    \draw[->,thick] (6.6,2.9) -- (6.6,2.1);
  % Incoming sequence
    \draw [fill=blue] (0,1) rectangle ++(1,1) node[midway] {};
     %Outgoing
\draw [fill=blue] (6.1,1) rectangle ++(1,1) node[midway] {};  
  % Big square with B^t_{i,j}
  \foreach \x in {0,...,2}
    \foreach \y in {0,...,2}
      \pgfmathtruncatemacro\i{3-\y}
      \pgfmathtruncatemacro\j{1+\x}
      \draw (2+\x,\y) rectangle ++(1,1) 
      node[midway] {};
% Color buffer	
	\draw[fill=green] (3,2) rectangle ++(1,1);  
	\draw[fill=blue] (4,2) rectangle ++(1,1);
	
	\draw[fill=lila] (2,1) rectangle ++(1,1); 
	\draw[fill=lila] (3,1) rectangle ++(1,1); 	
	\draw[fill=blue] (4,1) rectangle ++(1,1);      
     % Arrows
   \foreach \x in {0,...,2}
    \draw[->,thick] (1,1.5) -- (2,\x+0.5);  
  \foreach \x in {0,...,2}
    \draw[->,thick] (5,\x+0.5) -- (6,1.4+0.1*\x); 
\end{tikzpicture}
\vspace{0.2cm}
         \caption{Greedy retrieval action mask: Only retrieving from lane 1 or 2 is allowed.}
     \end{subfigure} \\
     
     \vspace{0.5cm}

     \begin{subfigure}[t]{0.47\textwidth}        
        \begin{tikzpicture}[scale=0.65]
\tikzstyle{every node}=[font=\footnotesize, align=center,execute at begin node=\setlength{\baselineskip}{11pt}]
% next incoming color
    \node[align=center] at (0.5,3.3) {color of next \\ incoming car};
    \draw[->,thick] (0.5,2.9) -- (0.5,2.1);
% current color
    \node[] at (6.6,3.3) {current color};
    \draw[->,thick] (6.6,2.9) -- (6.6,2.1);
  % Incoming sequence
    \draw [fill=blue] (0,1) rectangle ++(1,1) node[midway] {};
     %Outgoing
\draw [fill=blue] (6.1,1) rectangle ++(1,1) node[midway] {};  
  % Big square with B^t_{i,j}
  \foreach \x in {0,...,2}
    \foreach \y in {0,...,2}
      \pgfmathtruncatemacro\i{3-\y}
      \pgfmathtruncatemacro\j{1+\x}
      \draw (2+\x,\y) rectangle ++(1,1) 
      node[midway] {};
% Color buffer	
	\draw[fill=red] (3,2) rectangle ++(1,1);  
	\draw[fill=green] (4,2) rectangle ++(1,1);
%first lane
	\draw[fill=lila] (2,1) rectangle ++(1,1); 
	\draw[fill=lila] (3,1) rectangle ++(1,1); 	
	\draw[fill=red] (4,1) rectangle ++(1,1);      
     % Arrows
   \foreach \x in {0,...,2}
    \draw[->,thick] (1,1.5) -- (2,\x+0.5);  
  \foreach \x in {0,...,2}
    \draw[->,thick] (5,\x+0.5) -- (6,1.4+0.1*\x); 
\end{tikzpicture}
\vspace{0.2cm}
         \caption{Fast track action mask: Only storing in lane 3 is allowed.}
     \end{subfigure}
\hspace{0.5cm}     
     \begin{subfigure}[t]{0.47\textwidth}
         %\centering
         \begin{tikzpicture}[scale=0.65]
\tikzstyle{every node}=[font=\footnotesize, align=center,execute at begin node=\setlength{\baselineskip}{11pt}]
% next incoming color
    \node[align=center] at (0.5,3.3) {color of next \\ incoming car};
    \draw[->,thick] (0.5,2.9) -- (0.5,2.1);
% current color
    \node[] at (6.6,3.3) {current color};
    \draw[->,thick] (6.6,2.9) -- (6.6,2.1);
  % Incoming sequence
    \draw [fill=blue] (0,1) rectangle ++(1,1) node[midway] {};
     %Outgoing
\draw [fill=blue] (6.1,1) rectangle ++(1,1) node[midway] {};  
  % Big square with B^t_{i,j}
  \foreach \x in {0,...,2}
    \foreach \y in {0,...,2}
      \pgfmathtruncatemacro\i{3-\y}
      \pgfmathtruncatemacro\j{1+\x}
      \draw (2+\x,\y) rectangle ++(1,1) 
      node[midway] {};
% Color buffer	
	\draw[fill=blue] (3,2) rectangle ++(1,1);  
	\draw[fill=green] (4,2) rectangle ++(1,1);
	
	\draw[fill=blue] (3,1) rectangle ++(1,1); 
	\draw[fill=red] (4,1) rectangle ++(1,1);      
     % Arrows
   \foreach \x in {0,...,2}
    \draw[->,thick] (1,1.5) -- (2,\x+0.5);  
  \foreach \x in {0,...,2}
    \draw[->,thick] (5,\x+0.5) -- (6,1.4+0.1*\x); 
\end{tikzpicture}
\vspace{0.2cm}
     	\caption{Greedy storage action mask: Only storing in lane 1 or 2 is allowed.}
     \end{subfigure}
%\hspace{0.5cm}     
        \caption{Illustration of the four considered action masks.}
        \label{fig:Action_mask_illustration}
\end{figure}
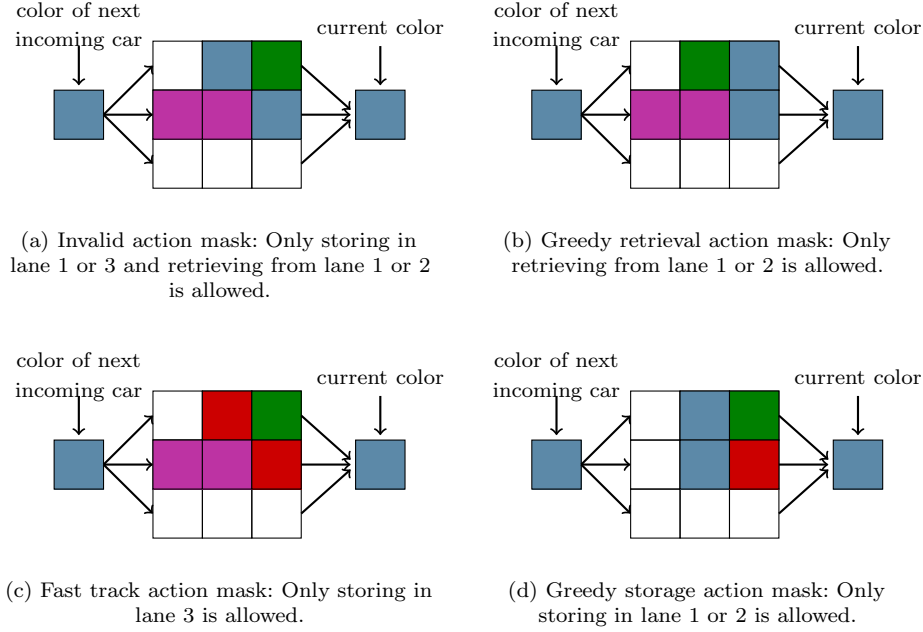

We consider four action masks as illustrated in \Cref{fig:Action_mask_illustration}. 
The first action mask $m^{\text{INV}}$ excludes invalid actions, i.e., storing in a full lane or retrieving from an empty lane
\begin{equation}
m^{\text{INV}}(a, s_t)=\begin{cases}
1&\text{if } a\in\{1,\dots,L\} \text{ and } B_{t,a,W}\neq 0 \\ 
1&\text{if } a\in\{L+1,\dots,2L\} \text{ and }  B_{t,a-L,1}=0 \text{ and } e_{t,1}\neq 0 \\
0& \text{else}.
\end{cases}
\end{equation} 
Adding this action mask hence waives the need to penalize invalid actions with negative rewards. Thus, the learning episodes can also be expected to be shorter, resulting in a more efficient learning process overall.

The next two action masks $m^{\text{GR}}$ and $m^{\text{FT}}$ enforce so called \textit{greedy retrieval} and \textit{fast-track}. These provably optimal actions are not possible in every state. We thus define two action masks that enforce these actions whenever they are possible, following the general procedure of enforcing optimal actions (see \Cref{sec:enforce_optimal}). % \Cref{eq:prescribe_optimal}.
An action is called \textit{greedy retrieval} if it retrieves a car from the rightmost position of a buffer lane that has the same color as the current color of the outgoing sequence. The greedy retrieval mask $m^{\text{GR}}(a, s_t)$ is given as
\begin{equation}
m^{\text{GR}}(a, s_t)=\begin{cases}
1 &\text{if } a\in\{1,\dots,L\} \text{ and } p_t=B_{t,a,W} \\
1 &\text{if } p_t\neq B_{t,i,W}\ \forall i=1,\dots, L \\
0 &\text{else}.
\end{cases}
\end{equation} 

We refer to \textit{fast-track} as storing a car in an empty buffer lane in order to directly retrieve in the next step without causing a color change.  
The fast-track mask $m^{\text{FT}}$ enforces store actions to empty lanes if a subsequent greedy retrieval from this lane is possible.
%, consistent with \Cref{thm:fast-track}.
Otherwise, the mask allows all actions if fast-track is not possible
\begin{equation}
m^{\text{FT}}(a, s_t)=\begin{cases}
1 &\text{if } a\in\{L+1,\dots,2L\} \text{ and } B_{t,a-L,1}=0 \text{ and } p_t=e_{t,1}\\
1 &\text{if } p_t\neq e_{t,1} \text{ or } B_{t,i,1}\neq 0, \ \forall i=1,\dots, L \\
0 &\text{else}.
\end{cases}
\end{equation}

%The mask enforces greedy retrieval if it is possible. If greedy retrieval is not possible, the mask allows all actions.

Finally, we consider \textit{greedy storage}, i.e., storing the current car from the incoming sequence in a buffer lane, where the leftmost car is of the same color. While we cannot prove that the heuristic is indeed optimal, it still presents a reasonable heuristic. However, our action space does not possess any structure that follows a reasonable order as colors are uniquely different. We therefore cannot follow  \Cref{eq:prescribe_approx_distance} or \Cref{eq:prescribe_approx_ordered} to approximately prescribe the heuristic. Nevertheless, we still  evaluate a fourth action mask $m^{\text{GS}}(a, s_t)$ that enforces greedy storage
\begin{equation}
m^{\text{GS}}(a, s_t)=\begin{cases}
1 &\text{if } a\in\{L+1,\dots,2L\} \text{ and } \exists j \in\{2,\dots,W\}\colon \\ 
  & e_{t,1}=B_{t,a-L,j}, B_{t,a-L,j-1}=0 \\
1 &\text{if } \nexists i \in\{1,\dots,L\}, j \in\{2,\dots,W\}\colon \\
  & e_{t,1}=B_{t,i,j}, B_{t,i,j-1}=0\\
0 &\text{else}.
\end{cases}
\end{equation}

%% note

Note that employing the aforementioned action masks does not solve the paint shop problem. In fact, greedy retrieval, greedy storage, and fast-track actions are often not possible. Therefore, an effective solution policy still needs to perform several optimal storage and retrieval actions.

%% combinations

So far, we considered several action masks independently. However, we can combine multiple action masks and perform our evaluations using the procedure for combining action masks defined in \Cref{eq:combine_add} and \Cref{eq:combine_seq}. We give the least priority to $m^{\text{GS}}$, as we do not know whether greedy storage is indeed optimal
\begin{itemize}
\item $m^{\text{INV}}\oplus  \left( \left(m^{\text{GR}} \ogreaterthan m^{\text{FT}} \right) \ogreaterthan m^{\text{GS}} \right)  $ (all action masks)
\item $m^{\text{INV}}\oplus \left(m^{\text{GR}} \ogreaterthan m^{\text{FT}} \right) $ (invalid + greedy retrieval + fast-track)
\item $m^{\text{INV}}\oplus m^{\text{GR}}$ (invalid + greedy retrieval)
\item $m^{\text{INV}}$ (invalid)
\end{itemize}

\vspace{-0.4cm}

\definecolor{ao}{rgb}{0.0, 0.0, 1.0}
\definecolor{armygreen}{rgb}{0.29, 0.33, 0.13}
\definecolor{antiquefuchsia}{rgb}{0.57, 0.36, 0.51}
\definecolor{bronze}{rgb}{0.8, 0.5, 0.2}
\definecolor{auburn}{rgb}{0.43, 0.21, 0.1}

\begin{figure}[H]
\footnotesize
\begin{subfigure}[b]{0.5\textwidth}
\centering
\pgfplotsset{every tick label/.append style={font=\scriptsize}}
     \centering
        \begin{tikzpicture}
		\begin{axis}[
			width=\textwidth,
		y label style={font=\footnotesize},
		x label style={font=\footnotesize},
		title style={font=\small},
		xtick={1, 2, 3, 4,5,6,7,8},
		ytick={10, 20, 30, 40,50,60,70,80, 90},
		ymax=95, ymin=0,
		  ylabel={Color changes},
	 		xlabel={Buffer size},
		   ymajorgrids,
		   axis lines*=left,
		  mark size=2pt,
		  line width=1pt,
		   legend style={
				anchor=north,
				font=\tiny,
				legend cell align=left,
				legend pos=south west,
				draw = none
			}
		]

		%% HIER KOMMT ES AUF DIE LABELS AN --> \ref
		%% SETZE AUF line width=1pt, damit es spaeter in der custom legend bold wird...
		
\addplot[color=black,mark = o] table [x=buffer_size, y=Max, col sep=comma] {ps_5colors.csv}; \label{pgfplots:main_Max}
		\addplot[color=armygreen,mark = o] table [x=buffer_size, y=FT_rep, col sep=comma] {ps_5colors.csv};\label{pgfplots:main_FT_rep}
		\addplot[color=antiquefuchsia,mark = o] table [x=buffer_size, y=GR_rep, col sep=comma] {ps_5colors.csv};\label{pgfplots:main_GR_rep}
		\addplot[color=ao,mark = o] table [x=buffer_size, y=INV_rep, col sep=comma] {ps_5colors.csv};\label{pgfplots:main_INV_rep}
		\addplot[color=green,mark = o] table [x=buffer_size, y=no_mask_repeated, col sep=comma] {ps_5colors.csv};\label{pgfplots:main_no_mask_rep}
		\addplot[color=red,mark = o] table [x=buffer_size, y=Greedy, col sep=comma] {ps_5colors.csv};\label{pgfplots:main_Greedy}
		\end{axis}
		\end{tikzpicture}
		\caption{\footnotesize Instances with 5 colors.}
\end{subfigure}
\begin{subfigure}[b]{0.5\textwidth}
\centering
\pgfplotsset{every tick label/.append style={font=\scriptsize}}
     \centering
        \begin{tikzpicture}
		\begin{axis}[
			width=\textwidth,
		y label style={font=\footnotesize},
		x label style={font=\footnotesize},
		title style={font=\small},
		xtick={1, 2, 3, 4,5,6,7,8},
		ytick={10, 20, 30, 40,50,60,70,80, 90},
		ymax=95, ymin=0,
	   %xmajorgrids,
		  ylabel={Color changes},
	 		xlabel={Buffer size},
		   ymajorgrids,
		   axis lines*=left,
		  mark size=2pt,
		  line width=1pt,
		   legend style={
				anchor=north,
				font=\tiny,
				legend cell align=left,
				legend pos=south west,
				draw = none
			}
		]
		\addplot[color=black,mark = o] table [x=buffer_size, y=Max, col sep=comma] {ps_10colors.csv};
		\addplot[color=armygreen,mark = o] table [x=buffer_size, y=FT_rep, col sep=comma] {ps_10colors.csv};
		\addplot[color=antiquefuchsia,mark = o] table [x=buffer_size, y=GR_rep, col sep=comma] {ps_10colors.csv};
		\addplot[color=ao,mark = o] table [x=buffer_size, y=INV_rep, col sep=comma] {ps_10colors.csv};
		\addplot[color=green,mark = o] table [x=buffer_size, y=no_mask_repeated, col sep=comma] {ps_10colors.csv};
		\addplot[color=red,mark = o] table [x=buffer_size, y=Greedy, col sep=comma] {ps_10colors.csv};
		\end{axis}
		\end{tikzpicture}
		\caption{\footnotesize Instances with 10 colors.}
\end{subfigure}
\vspace{0.2cm}

%% 15 colors in der Mitte

\centering
\begin{subfigure}[b]{0.5\textwidth}
\centering
\pgfplotsset{every tick label/.append style={font=\scriptsize}}
     \centering
        \begin{tikzpicture}
		\begin{axis}[
			width=\textwidth,
		y label style={font=\footnotesize},
		x label style={font=\footnotesize},
		title style={font=\small},
		xtick={1, 2, 3, 4,5,6,7,8},
		ytick={10, 20, 30, 40,50,60,70,80, 90},
		ymax=95, ymin=0,
	   %xmajorgrids,
		  ylabel={Color changes},
	 		xlabel={Buffer size},
		   ymajorgrids,
		   axis lines*=left,
		  mark size=2pt,
		  line width=1pt,
		   legend style={
				anchor=north,
				font=\tiny,
				legend cell align=left,
				legend pos=south west,
				draw = none
			}
		]
		\addplot[color=black,mark = o] table [x=buffer_size, y=Max, col sep=comma] {ps_15colors.csv};
		\addplot[color=armygreen,mark = o] table [x=buffer_size, y=FT_rep, col sep=comma] {ps_15colors.csv};
		\addplot[color=antiquefuchsia,mark = o] table [x=buffer_size, y=GR_rep, col sep=comma] {ps_15colors.csv};
		\addplot[color=ao,mark = o] table [x=buffer_size, y=INV_rep, col sep=comma] {ps_15colors.csv};
		\addplot[color=green,mark = o] table [x=buffer_size, y=no_mask_repeated, col sep=comma] {ps_15colors.csv};
		\addplot[color=red,mark = o] table [x=buffer_size, y=Greedy, col sep=comma] {ps_15colors.csv};
		\end{axis}
		\end{tikzpicture}
		\caption{\footnotesize Instances with 15 colors.}
\end{subfigure} \\
\vspace{0.2cm}

%% Legend for all
\begin{tikzpicture}
	\matrix[
	 font=\footnotesize,
    matrix of nodes,
    anchor=west,
    draw,% Rahmen um Legende
    inner sep=0.2em,
    line width=1pt,
    cells={line width=1pt, anchor=west}
  ]
  at(0.6,2.2){
    \ref{pgfplots:main_Max} & RL (all masks) &
    \ref{pgfplots:main_FT_rep} & RL (invalid + greedy retrieval + fast-track mask) \\
\ref{pgfplots:main_INV_rep} & RL (invalid mask) &    
    \ref{pgfplots:main_GR_rep} & RL (invalid + greedy retrieval mask) \\
    \ref{pgfplots:main_no_mask_rep} & RL (no mask) &
    \ref{pgfplots:main_Greedy} & Greedy heuristic \\
   };
    \end{tikzpicture}
\caption{Evaluation results (color changes) for all RL approaches and Greedy heuristic.\label{fig:actionmasking_comparison_plots}}
\end{figure}

\subsection{Results}

We compare the number of color changes for RL combined with all action masks against the greedy heuristic. This heuristic simply applies the actions suggested by greedy retrieval, fast-track, and greedy storage. If none of these are possible, it performs a random possible action. 
We vary the buffer size from 2x2 to 8x8 and number of colors as $C=5,10,15$. For each buffer size, we generate ten random incoming sequences of length 100. Each sequence is randomly generated by sampling the color at each sequence position independently and with equal probability.

%% results

The results are shown in \Cref{fig:actionmasking_comparison_plots}. 
We find that combining a greater number of action masks generally decreases the number of color changes.
Furthermore, we observe that this improvement depends on the buffer size.
For small 2x2 buffers, the benefit is almost negligible, while the performance increase is more significant for larger buffers.
For 8x8 buffers in particular, RL with the combination of all action masks causes approximately half the number of color changes compared to RL without any action masking.

\begin{figure}[H]
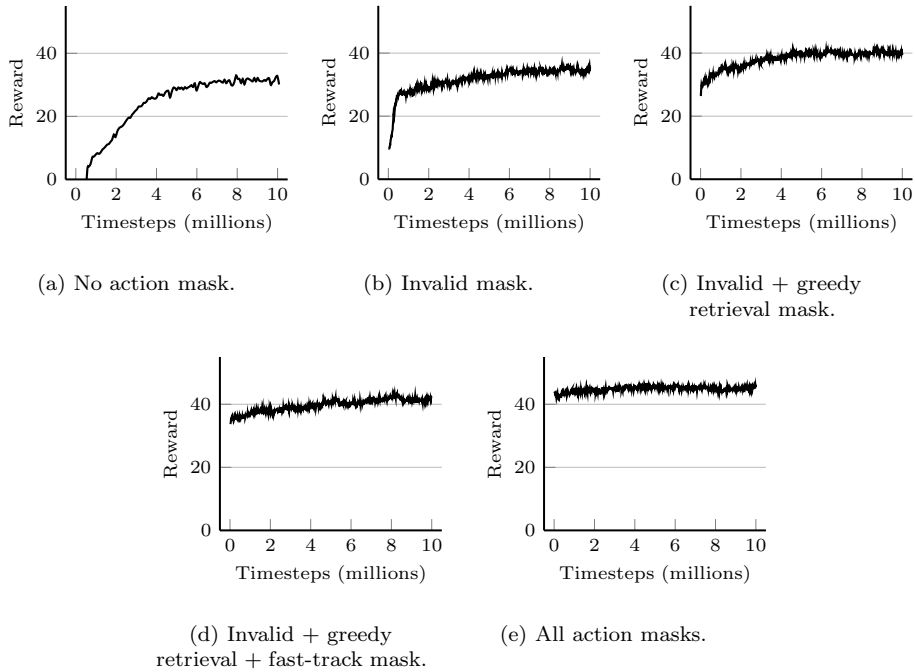

\centering
\footnotesize
\captionsetup[subfigure]{justification=centering,font=footnotesize,skip=0pt}
     \begin{subfigure}[t]{0.3\textwidth}
		\input{line_plot_template_PS.tex}
		\addplot[color=black,line width=0.8pt] table [x=Step, y=Value, col sep=comma] {10_4_100_0_False_5_bitmap_1.csv};
		\end{axis}
		\end{tikzpicture}
\vspace{0.1cm}
		\caption{No action mask.}
     \end{subfigure}
   \hfill
     \begin{subfigure}[t]{0.3\textwidth}
		\input{line_plot_template_PS.tex}
		\addplot[color=black,line width=0.8pt] table [x=Step, y=Value, col sep=comma] {10_4_100_0_False_5bitmap_actionmask_1.csv};
		\end{axis}
		\end{tikzpicture}         
    \vspace{0.1cm}
        \caption{Invalid mask.}
        
     \end{subfigure}
   	\hfill
     \begin{subfigure}[t]{0.3\textwidth}
		\input{line_plot_template_PS.tex}
		\addplot[color=black,line width=0.8pt] table [x=Step, y=Value, col sep=comma] {10_4_100_0_False_5bitmap_actionmask_enforcegreedy_42_1.csv};
		\end{axis}
		\end{tikzpicture}   
		\vspace{0.1cm}      
        \caption{Invalid + greedy retrieval mask.}
     \end{subfigure}
   	\vspace{0.4cm}
     
     %% 2nd row
          \begin{subfigure}[b]{0.3\textwidth}
		\input{line_plot_template_PS.tex}
		\addplot[color=black,line width=0.8pt] table [x=Step, y=Value, col sep=comma] {10_4_100_0_False_5bitmap_actionmask_enforcegreedy_ft_42_1.csv};
		\end{axis}
		\end{tikzpicture}
\vspace{0.1cm}
		\caption{Invalid + greedy retrieval + fast-track mask.}
     \end{subfigure}
   \hspace{0.5cm}
     \begin{subfigure}[b]{0.3\textwidth}
		\input{line_plot_template_PS.tex}
		\addplot[color=black,line width=0.8pt] table [x=Step, y=Value, col sep=comma] {10_4_100_0_False_5bitmap_actionmask_enforcegreedy_sar_ft_42_1.csv};
		\end{axis}
		\end{tikzpicture}         
\vspace{0.1cm}        
        \caption{All action masks. \newline}
     \end{subfigure}

        \caption{Learning curves for RL models with 10 colors and 4x4 buffer and varying action masks.}
        \label{fig:Learning_curves_PS}
\end{figure}

We also analyze the learning curves for a problem instance with 10 colors, 4x4 buffer, and all evaluated combinations of action masks, as shown in \Cref{fig:Learning_curves_PS}.
The plots show that the RL policies trained with more action masks converge to a greater reward. 
Furthermore, the learning process is considerably faster if more action masks are added to policy learning. Specifically, the model without action masking only reaches positive rewards after 500,000 time steps. Conversely, including the invalid mask ensures positive rewards from the first learning episode. Increasing the number of action masks leads to a higher initial reward over the first learning episodes. Therefore, action masking not only leads to higher overall rewards but also mitigates the ``cold start'' problem of an untrained RL policy.

\section{Problem 2: Peak Load Management}
\label{sec:loadmanagement}

A peak load management system (LMS) needs to decide when to turn off electric devices with high energy consumption to keep the peak load below a given threshold \citep{Papier.2016,Dong.2017}. Violations of this threshold must be avoided at any time point as they entail large payments to the energy supplier. 

\subsection{Environment}

The company can turn off large energy consumers like air conditioning to reduce the peak load. Hence, the action space consists of only two actions: turning off air conditioning and leaving it on
\begin{equation}
A^{\text{LMS}} = \lbrace \text{off, on} \rbrace.\label{eq:actions_lms}
\end{equation}
If the air conditioning is turned off, energy consumption is considerably reduced during the next time period. 

We consider a time horizon of one day, while one period lasts 15 minutes. This yields a finite time horizon of $T=24 \times 4 = 96$ time steps. However, due to legal requirements, executing action \texttt{off} to turn off air conditioning is constrained to a total of three executions per day. Therefore, action \texttt{off} should only be performed when a high peak load occurs.

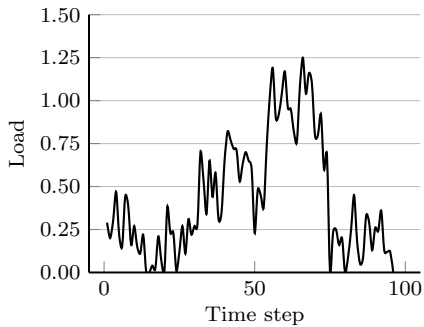
\begin{figure}[H]
\centering

		\begin{tikzpicture}
		\begin{axis}[
		   xmin=0, xmax=100,
			ymin=0, ymax=1.5,
			ytick={0,0.25,0.5,0.75,1,1.25,1.5},
			width=0.5\textwidth,
			height=5cm,
			yticklabel=\pgfkeys{/pgf/number format/.cd,fixed,precision=2,zerofill}\pgfmathprintnumber{\tick},
			tick label style={font=\footnotesize},
		y label style={font=\footnotesize},
		x label style={font=\footnotesize, at={(axis description cs:0.5,-0.1)},anchor=north},
		title style={font=\small},
		   %xmajorgrids,
		   enlarge x limits=0.05,
		   ylabel={Load},
	 		xlabel={Time step},
		   ymajorgrids,
		   axis lines*=left,
		  mark size=0pt,
		   legend style={
				at={(0.5,-0.2)},
				anchor=north,
				legend columns=2,
				font=\footnotesize,
				/tikz/every even column/.append style={column sep=0.5cm}
			},
		    smooth
		]
		%% mean
		\addplot[color=black,line width=0.8pt] table [x=x, y=true_load, col sep=comma] {load_tikz.csv};

		\end{axis}
		\end{tikzpicture}
		\caption{Load curve for load management system over 96 timesteps.\label{fig:lms}}
\end{figure}

The system's load curve is illustrated in \Cref{fig:lms}. While this curve remains constant throughout the study, it is initially unknown to the reinforcement learning policy and must be discovered during training. We do not evaluate different or changing curves, as the setup with a fixed curve and noisy forecasts already creates sufficient complexity for analyzing the policy's learning capabilities.

%% state

The state of the LMS is given by the actual consumption $c_{t-1}$ of the last time period, the estimated consumption $\widehat{c}_t$ of the next time period, and how often the action \texttt{off} can still be performed, denoted by $n$
\begin{equation}
s_t^{\text{LMS}} = (c_{t-1}, \widehat{c}_t, n).\label{eq:state_lms}
\end{equation}
The estimated consumption is given by a predictive model $\widehat{c}_t= c_t + \varepsilon$ with normally distributed error $\varepsilon\sim\mathcal{N}(0, \sigma)$ and standard deviation $\sigma$. 

%% reward

The goal of the LMS problem is to keep the maximum load over all time steps below a given threshold $\zeta$. The RL policy's actions to reduce the peak load are successful if the maximum load $c^*=\max_{t=0}^{T-1}\{c_t\mid a_t=\text{on}\}$ over all $T=96$ time steps is lower than the peak load threshold of $\zeta=1.24$. Therefore, whether or not the RL policy was successful in managing the peak load can only be determined at the end of the time horizon. 
Accordingly, we define the following reward function 
\begin{equation}
R^{\text{LMS}}(a,s_t) = \begin{cases} 
\hfill 1 &\text{if}\ t=T-1 \land c^*<\zeta \\
\hfill -1 &\text{if}\ t=T-1 \land c^*\geq\zeta \\
\hfill 0 &\text{if}\ t<T-1 . \end{cases} \label{eq:reward_lb}
\end{equation}

\subsection{Action Masking}

We consider action masks that restrict the execution of action \texttt{off}. Given that \texttt{off} can only be performed a limited number of times due to legal constraints, it seems reasonable to perform it only, if the estimate of the next period $\widehat{c}_t$ is sufficiently high. Therefore, we define the mask with respect to $\widehat{c}_t$, so that \texttt{off} can only be performed if the predicted consumption is above a given threshold $\theta^{LMS}$ 
\begin{equation}
m(a, s_t) = \begin{cases} 
1 &\text{if}\ \widehat{c}_t \geq \theta^{LMS} \text{ and } a=\text{off} \\
1 &\text{if}\ a=\text{on} \\
0 &\text{else}. \end{cases} \label{eq:intent_LB}
\end{equation}
We consider multiple masks with varying $\theta^{LMS}$ between 0 and 1.20. Setting $\theta^{LMS}=0$ corresponds to excluding action masking. The action masks become more restrictive for higher values of $\theta^{LMS}$.

\subsection{Results}

In our evaluation, we vary the parameters $\sigma$ and $\theta^{LMS}$. 
For each parameter setting, we train an RL policy for one million time steps and then report the number of successfully solved instances out of 100 evaluations to account for randomness in $\widehat{c}_t$. The results are presented in \Cref{tbl:LMS}. We present the fraction of instances that could be successfully solved (i.e., the peak load always remained below the threshold $\zeta$). 
We find that RL without action masking (see column ``none'') is not able to learn an effective policy, irrespective of the noise level. The same results can be observed for the lower values $\theta^{LMS}=0.20, \ 0.40, \ 0.60$. However, if $\theta^{LMS}\geq 0.80$, the trained policies manage to solve considerable fractions of the problem instances, depending on the noise level. Note that apart from the setting with perfect load forecasts (noise level=none, $\widehat{c}_t=c_t$), we do not expect to consistently succeed due the randomness of $\widehat{c}_t$ and the general difficulty of the problem. 

\vspace{0.2cm}

\begin{figure}[H]
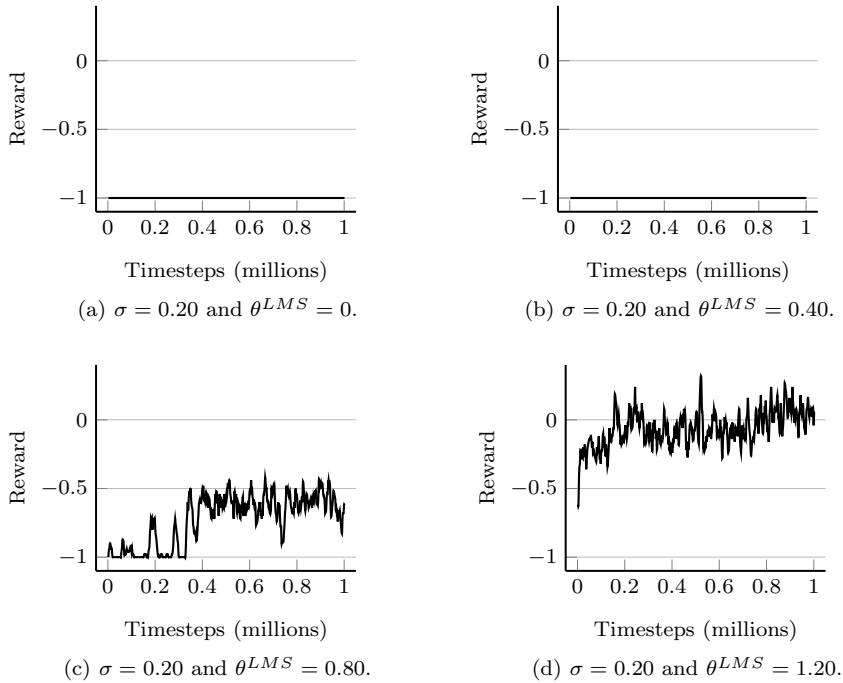

\centering
%\footnotesize
\captionsetup[subfigure]{justification=centering,font=footnotesize,skip=0pt}
     \begin{subfigure}[b]{0.48\textwidth}
		\input{line_plot_template.tex}
		\addplot[color=black,line width=0.8pt] table [x=Step, y=Value, col sep=comma] {LMS_0.0_0.2.csv};
		\end{axis}
		\end{tikzpicture}
			\vspace{0.1cm}
		\caption{$\sigma = 0.20$ and $\theta^{LMS}=0$.}
     \end{subfigure}
  % \hspace{0.5cm}
  \hfill
     \begin{subfigure}[b]{0.48\textwidth}
		\input{line_plot_template.tex}
		\addplot[color=black,line width=0.8pt] table [x=Step, y=Value, col sep=comma] {LMS_0.0_0.2.csv};
		\end{axis}
		\end{tikzpicture}  
		\vspace{0.1cm}       
        \caption{$\sigma = 0.20$ and $\theta^{LMS}=0.40$.}
     \end{subfigure}
   	\vspace{0.5cm}
     
          \begin{subfigure}[b]{0.48\textwidth}
		\input{line_plot_template.tex}
		\addplot[color=black,line width=0.8pt] table [x=Step, y=Value, col sep=comma] {LMS_0.8_0.2.csv};
		\end{axis}
		\end{tikzpicture}
		\vspace{0.1cm}
		\caption{$\sigma = 0.20$ and $\theta^{LMS}=0.80$.}
     \end{subfigure}
   \hfill
     \begin{subfigure}[b]{0.48\textwidth}
		\input{line_plot_template.tex}
		\addplot[color=black,line width=0.8pt] table [x=Step, y=Value, col sep=comma] {LMS_1.2_0.2.csv};
		\end{axis}
		\end{tikzpicture}
		\vspace{0.1cm}   
        \caption{$\sigma = 0.20$ and $\theta^{LMS}=1.20$.}
     \end{subfigure}
     
        \caption{Learning curves for RL models with $\sigma = 0.20$ and thresholds $\theta^{LMS}=$ 0, 0.40, 0.80, 1.20.}
        \label{fig:Learning_curves_LMS}
\end{figure}

%% learning curves

We also consider the learning curves for the noise level $\sigma = 0.20$ shown in \Cref{fig:Learning_curves_LMS}.
The RL models with $\theta^{LMS}=0.00$ and $\theta^{LMS}=0.40$ are not able to learn an effective policy.
However, the RL models with $\theta^{LMS}=0.80$ and $\theta^{LMS}=1.20$ converge to an average reward of roughly $-0.50$ and a slightly positive reward, respectively. Note that a reward of $-0.50$ corresponds to solving 25 \% of the problem instances.

\begin{table}[H]
\setlength{\tabcolsep}{3pt}
\caption{Fraction of solved instances from 100 episodes after policy learning for one million time steps for different action mask thresholds and noise levels. \label{tbl:LMS}}

\footnotesize
\centering
%\begin{tabular}{l*{7}{S[table-format=1.2,table-column-width=1.3cm, parse-numbers=false]}}
\begin{tabular}{c*{7}{R{1.2cm}}}
\toprule
& \multicolumn{7}{c}{action mask threshold $\theta^{LMS}$} \\
\cmidrule(lr){2-8}
noise level $\sigma$ & {none} & {0.20} & {0.40} &  {0.60} &   {0.80} &   {1.00} &   {1.20} \\
\midrule
none         & 0 \% & 0 \% & 0 \% & 0 \% & 56 \% & 100 \% & 100 \% \\
0.10         & 0 \% & 0 \% & 0 \% & 0 \% & 46 \% &  74 \% &  85 \% \\
0.20         & 0 \% & 0 \% & 0 \% & 0 \% & 23 \% &  31 \% &  61 \% \\
0.30 		 & 0 \% & 0 \% & 0 \% & 0 \% & 17 \% &  23 \% &  48 \% \\
0.40         & 0 \% & 0 \% & 0 \% & 0 \% &  0 \% &  17 \% &  34 \% \\
\bottomrule
\end{tabular}
\end{table}

%\vspace{-0.5cm}

\section{Problem 3: Inventory Management}
\label{sec:inventorymanagement}

%% allgemein inventory

The objective of the inventory management problem is to minimize the cost of operating an inventory \citep[e.g.,][]{Boute.2022,Heger.2024}. At each time step, a stochastic demand is realized and the operating agent places an order to replenish the inventory. However, orders do not arrive instantly as they are delayed by the lead time, which can be either stochastic or deterministic.
The total costs of operating the inventory are the sum of inventory and lost sales costs.\footnote{We consider the lost sales setting instead of backlogged systems since it is more challenging and interesting. In fact, the so called base-stock policy is proven to be optimal for backlogged systems \citep{zipkin2008old}.}

\begin{table}[H]
\footnotesize
\caption{Overview of parameters and variables.\label{tbl:parameters_inventory}}
\centering

\begin{tabular}{L{2cm}p{9cm}}
\toprule
\textbf{Parameters} \\
\midrule
$c=1$ & Holding cost\\
$p\in \{1,4\}$ & Lost sales cost\\
$N\in \{4,8\}$ & Maximum lead time: $N=4$ (deterministic) or $N=8$ (stochastic)\\
$\lambda=5$ & Mean demand of Poisson distribution \\
$H=5000$ & Time horizon \\
$\Delta=10$ & Discretization value of action space\\
\midrule
\textbf{Variables}\\
\midrule
$I_t$ & Inventory at time $t$ \\
$Q_{t,i}$ & Pipeline vector at time $t$ and position $i\in\{1,\dots N\}$ \\
$d_t$ & Demand at time $t$ \\
$L_t$ & Lead time  at time $t$: deterministic ($L_t=N$) or stochastic  $L_t\sim\text{Unif}(1,N)$\\
\bottomrule
\end{tabular}
\end{table}

In addition, there is a trade-off between minimizing losing sales and minimizing inventory holding costs. One major challenge of the inventory problem is to learn the distribution of demand and lead times. However, due to its intensive study, several useful heuristics were developed in prior works \citep{zipkin2008old, perera2023survey}, which we consider for action masking.

\subsection{Environment}

We adopt the RL formalization of the inventory management problem by \citet{Gijsbrechts.2022} and adjust it to include stochastic lead times. An overview of all parameters and variables is provided in \Cref{tbl:parameters_inventory}. 
The state of the system is given by the current inventory and the pipeline vector of incoming orders
\begin{equation}
s^{\text{IM}}_t = (\underbrace{I_t}_{\substack{\text{Current}\\\text{inventory}}}, \underbrace{Q_{t,1},\dots Q_{t, N}}_{\substack{ \text{Pipeline vector of}\\ \text{incoming orders}}}).
\end{equation}
The pipeline vector has length $N$, denoting the maximum lead time.

The actions specify the amount of items to be reordered. We discretize the action space by only allowing quantities given as multiples of $\Delta=10$ 
\begin{equation}
A^{\text{IM}} = \lbrace 0, \Delta, 2\Delta, \dots, 10\Delta \rbrace.\label{eq:action_space_inventory}
\end{equation}

After a certain quantity is ordered, the next incoming order is added to the inventory level and the orders of the pipeline vector are shifted one step forward
\begin{equation}
I_t' = I_t+Q_{t,1}, \quad Q_{t,i}' = Q_{t,i+1}, \quad Q_{t,N}=0.
\end{equation}
Next, the demand $d_t$ is drawn from a Poisson distribution $\text{Pois}(\lambda)$ and subtracted from the inventory level. 
If the demand is greater than the inventory level, the inventory is set to zero and the remaining demand causes lost sales costs 
\begin{equation}
I_{t+1} = (I_t+Q_{t,1}-d_t)^+.\label{eq:inventory_basic}
\end{equation}

Lastly, the selected action $a$ is executed by sampling a lead time $L_t$ from the discrete uniform distribution and adding the ordered quantity $a$ to the pipeline vector at the respective lead time. We also evaluate deterministic lead times with $L_t=N$
\begin{align}
L_t &\sim \text{Unif}(1,N) \\
Q_{t+1,i} &= \begin{cases}
Q_{t,i}'+a  &\text{if } i=L_t\\
Q_{t,i}' &\text{else}.\\
\end{cases}
\end{align}

The reward function depends on whether or not sales are lost. 
If orders cannot be met, the agent receives a penalty proportionate to the number of lost sales.
Otherwise, holding costs are charged proportionate to the inventory
\begin{equation}
R^{\text{IM}}(a,s_t) = \begin{cases} 
-c (I_t+Q_{t,1}-d_t) &\text{if } I_t+Q_{t,1}-d_t\geq 0 \\
p (I_t+Q_{t,1}-d_t) &\text{else}.
\end{cases}
\end{equation}
Therefore, maximizing the cumulative reward over the given time horizon $H$ is equivalent to minimizing the total cost.

\subsection{Action Masking}

Our action masks are informed by the \textit{base stock heuristic} that was frequently used in prior studies \citep[e.g.,][]{zipkin2008old, huh2009asymptotic, Bijvank.2011}.
This heuristic depends on a parameter $S$, the \textit{base-stock level} and it chooses the action $h(s_t)=S-I_t-\sum_i Q_{t,i}$, i.e., the action that sets the sum of inventory and pipeline to the base-stock level.
The heuristic is employed independently of the lost sales costs $p$. However, for $p\rightarrow \infty$ it converges to the optimal policy \citep{huh2009asymptotic}. Consequently, we primarily consider smaller values of $p$. 

We derive two action masks based on this heuristic, $m^{\text{INT}}$ and $m^{\text{THR}}$.
The first action mask $m^{\text{INT}}$ allows actions that order a similar quantity as the prescribed action 
\begin{equation}
m^{\text{INT}}(a, s_t)=
\begin{cases}
1 &\text{if } |a-(S-I_t-\sum_i Q_{t,i})|\leq \Delta \\
0 &\text{else.}
\end{cases}
\end{equation}

The second action mask $m^{\text{THR}}$ uses the prescribed action $h(s_t)$ as a threshold and allows all actions that order at least as much
\begin{equation}
m^{\text{THR}}(a, s_t)=
\begin{cases}
1 &\text{if } a+I_t+\sum_i Q_{t,i} \geq S \\
0 &\text{else.}
\end{cases}
\end{equation}

For a given instance of the inventory problem, the optimal base-stock level $S$ can be determined by a search algorithm.
The optimal values have already been computed for many relevant situations and can be found in several sources \citep[e.g.,][]{huh2009asymptotic}. In our setting, the optimal values are given as $S=18$ for $p=1$ and $S=25$ for $p=4$.

\subsection{Results}

We compare the average inventory cost for the three action masks $m^{\text{INT}}$, $m^{\text{THR}}$ and no action mask.
We consider the scenarios of deterministic/stochastic lead times and lost sales costs of $p=1$ and $p=4$ with fixed holding costs $c=1$.
For each scenario, we train an RL policy for 1 million time steps and average its inventory cost over 100 evaluations. 
The time horizon for each evaluation is set to $H=5000$. 

\Cref{tbl:INV} provides the results.
For $p=1$, we observe that the RL policies with no action masking actually provide the lowest average inventory cost. However, if the lost sales costs increase, the RL policies with action masking yield lower inventory costs. This finding holds for both deterministic and stochastic lead times. In summary, we note that action masking using heuristics generally improves the resulting performances. Yet, as the \textit{base-stock level} heuristic is not optimal for $p=1$, we observe that omitting action masking leads to lower inventory costs.

\begin{table}[H]
\setlength{\tabcolsep}{3pt}
\footnotesize
\caption{Results of inventory problem. \label{tbl:INV}}

\centering
\begin{tabular}{cccC{1.5cm}}
\toprule
 Lost sales cost $p$ &  Lead time & Action mask  &  Avg.~costs \\
\midrule
               1 &      Deterministic &                   none &    \B  2.112 \\
               1 &      Deterministic & $m^{\text{INT}}$ &         2.186 \\
               1 &      Deterministic &          $m^{\text{THR}}$ &         2.198 \\
\midrule
               1 &     Stochastic &                   none &   \B   3.266 \\
               1 &     Stochastic & $m^{\text{INT}}$ &         3.560 \\
               1 &     Stochastic &          $m^{\text{THR}}$ &         3.791 \\
\midrule
               4 &      Deterministic &                   none &         5.330 \\
               4 &      Deterministic & $m^{\text{INT}}$ &  \B      5.058 \\
               4 &      Deterministic &          $m^{\text{THR}}$ &         5.155 \\

\midrule
               4 &     Stochastic &                   none &         8.311 \\
               4 &     Stochastic & $m^{\text{INT}}$ &    \B  7.933 \\
               4 &     Stochastic &          $m^{\text{THR}}$ &         8.105 \\

\bottomrule
\end{tabular}
\end{table}

\begin{figure}[h]
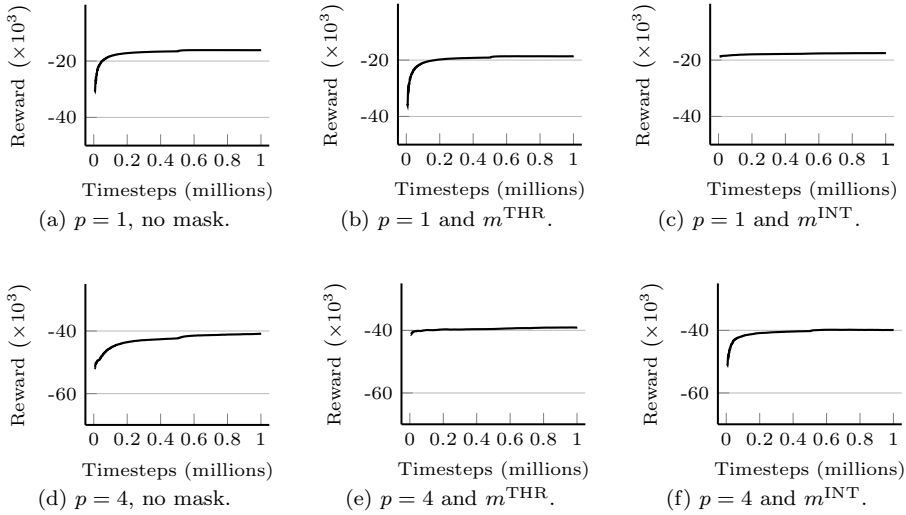

\centering
\footnotesize
\captionsetup[subfigure]{justification=centering,font=footnotesize,skip=0pt}
     \begin{subfigure}[b]{0.3\textwidth}
		\input{line_plot_template_INV.tex}
		\addplot[color=black,line width=0.8pt] table [x=Step, y=Value, col sep=comma] {1_no_poisson.csv};
		\end{axis}
		\end{tikzpicture}
		\caption{$p=1$, no mask.}
     \end{subfigure}
\hfill
     \begin{subfigure}[b]{0.3\textwidth}
		\input{line_plot_template_INV.tex}
		\addplot[color=black,line width=0.8pt] table [x=Step, y=Value, col sep=comma] {1_no_up-to-vector_poisson.csv};
		\end{axis}
		\end{tikzpicture}         
        \caption{$p=1$ and $m^{\text{THR}}$.}
     \end{subfigure}
   \hfill
     \begin{subfigure}[b]{0.3\textwidth}
		\input{line_plot_template_INV.tex}
		\addplot[color=black,line width=0.8pt] table [x=Step, y=Value, col sep=comma] {1_up-to-vector_bounded_poisson.csv};
		\end{axis}
		\end{tikzpicture}         
        \caption{$p=1$ and $m^{\text{INT}}$.}
     \end{subfigure}
\vspace{0.5cm}     

          \begin{subfigure}[b]{0.3\textwidth}
		\input{line_plot_template_INV_down.tex}
		\addplot[color=black,line width=0.8pt] table [x=Step, y=Value, col sep=comma] {4_no_poisson.csv};
		\end{axis}
		\end{tikzpicture}
		\caption{$p=4$, no mask.}
     \end{subfigure}
   \hfill
     \begin{subfigure}[b]{0.3\textwidth}
		\input{line_plot_template_INV_down.tex}
		\addplot[color=black,line width=0.8pt] table [x=Step, y=Value, col sep=comma] {4_no_up-to-vector_poisson.csv};
		\end{axis}
		\end{tikzpicture}         
        \caption{$p=4$ and $m^{\text{THR}}$.}
     \end{subfigure}
   \hfill
     \begin{subfigure}[b]{0.3\textwidth}
		\input{line_plot_template_INV_down.tex}
		\addplot[color=black,line width=0.8pt] table [x=Step, y=Value, col sep=comma] {4_up-to-vector_bounded_poisson.csv};
		\end{axis}
		\end{tikzpicture}         
        \caption{$p=4$ and $m^{\text{INT}}$.}
     \end{subfigure}
        \caption{Learning curves for RL models with stochastic lead times, $p=1$ (first row), $p=4$ (second row).}
        \label{fig:Learning_curves_INV}
\end{figure}

We also present the learning curves for stochastic lead times in \Cref{fig:Learning_curves_INV}.
The first row shows models with low lost sales cost ($p=1$).
We observe that the RL model without action masking smoothly reaches the highest reward, outperforming the models with action mask.
For the higher lost sales costs ($p=4$) shown in the second row, we find the opposite results, namely, that action masking is necessary to achieve a reward of roughly $-$40.000, corresponding to an average inventory cost of eight.

%% discussion learning curves

Note that the learning curves for the inventory optimization problem tend to be much smoother compared to those for the paint shop and load management problems.
This can be attributed to the more predictable dynamics of the inventory optimization problem.
In the load management problem, only a single non-zero reward is given at the end of each episode, and its value reflects the cumulative effect of all previous actions.
A small change in any action can have a significant impact on the final reward, leading to more fluctuations in the reward. 
Similarly, in the paint shop problem, individual actions can have persistent long-term effects; for example, a poor storage operation can block a buffer lane for many timesteps with detrimental effects on subsequent actions. 
By contrast, in the inventory management problem, the total reward is more gradually affected by individual actions. For instance, ordering 70 units instead of 60 causes only a small change to the total reward over the entire time horizon, leading to a smoother reward function and, consequently, smoother learning curves.

\section{Conclusion}
\label{sec:discussion}

%% summary of findings: 1, 2, 3 gut aber aufpassen...

We proposed the integration of human domain knowledge into reinforcement learning via action masking. Based on three OR problems, we showed that in addition to excluding invalid actions, action masking can also be employed to (i)~prescribe heuristics and (ii)~enforce provably optimal actions. In particular, our results suggest that the training process and the resulting performance of the trained policy can be improved considerably. Given that DRL does not provide any guarantees regarding convergence to a global optimum \citep{Sutton.2018}, employing action masking to enforce provably optimal actions in certain states led to considerable performance increases. For problems with constrained action spaces, action masking can even be necessary to learn effective policies that manage to solve a problem. Action masking can thus help to disallow unreasonable executions of the constrained action. As such, action masking presents an important ingredient for a successful implementation of RL in real-world operations. However, at the same time, we found that action masking might also decrease the performance if prescribing overly strict non-optimal heuristics, which prevent the learning of optimal policies.

%\subsection{Implications for Theory}

%% RL in OM generell

Our study adds to the literature that focuses on solving OR problems with RL. Specifically, we propose action masking as a method to integrate human domain knowledge to increase trust and the resulting solution quality. So far, related works employed action masking to disallow invalid actions \citep[e.g.,][]{Huang.2020,Luo.2022,Wang.2024}. We go beyond excluding invalid actions by guiding the policy towards near-optimal heuristic actions. %However, we do not strictly enforce heuristic actions to allow flexibility in the learning process.
From a theoretical view, a sufficient amount of training time should yield the same policy despite the lack of action masking. However, we found considerable differences in the solution quality between policies with and without action masking, even after millions of timesteps. While a randomly initialized policy first needs to learn the basic rules of the environment, action masking led to superior performance particularly in the early stages of learning. Action masking ensures that the policy follows human domain knowledge from the first learning episode. Although problems from daily operations are often considered as low-stake decision problems \citep{Notz.2024}, ensuring that AI models always operate as intended increases trust and user acceptance, which ultimately results in greater usage and reliance on the model \citep{Lehmann.2022}. 

%% reward shaping vs action masking

An alternative to action masking is given by reward shaping \citep{Cheng.2021}. Here, the reward function can be defined in a way that guides the RL policy towards heuristic actions. Reward shaping is less restrictive regarding how the policy evolves. An action receiving a large positive reward might not even be chosen if the policy never explores it. In addition, the higher reward for an action might not reach an overall higher cumulative reward over the entire learning episode. Conversely, action masking directly enforces the suggested action(s) in specific states, which restricts the policy from exploring potential alternatives. We therefore encourage future studies to look deeper into the pros and cons of action masking vs. reward shaping.

%\subsection{Implications for Practice}

%% User acceptance, trust, Safety guarantees

The use of action masks can enhance the safety and trustworthiness of RL systems in real-world applications. By constraining the agent's actions to safe and permissible options, organizations can ensure that the system operates within predefined boundaries, reducing the risk of unintended or harmful behaviors \citep{Notz.2024}. This is particularly important in safety-critical domains, such as autonomous driving, banking, and healthcare, where trustworthiness and reliability are paramount. In addition, employing action masking can lead to faster convergence and reduced training times, ultimately resulting in quicker implementation and deployment of RL-based solutions.

%% posthoc anpassungen

Furthermore, businesses can rely on action masking for post hoc policy adjustments. If human experts identify non-optimal actions that should be included or excluded, they simply need to define a novel action mask and apply it to a trained policy instead of retraining the policy. Businesses could also define separate action masks for different departments, products, and factories, further increasing the utility of a trained RL policy.

%\subsection{Limitations and Future Research}

Action masking presents a promising approach to embedding heuristics within RL systems.
However, it is not without limitations. One significant challenge arises when the heuristics themselves are imperfect or incomplete. In such cases, action masks may inadvertently constrain the agent too much, preventing it from exploring potentially better actions outside the scope of the provided heuristics. This can lead to suboptimal performance as we saw in the case of the base stock policy for inventory optimization with low lost sales costs. Future research should explore methods to dynamically adjust or refine action masks as the agent learns, ensuring that the masks do not overly restrict the agent's exploration while still guiding it effectively.

Furthermore, the comparison between action masking and traditional reward shaping highlights another area for further investigation. While action masking provides a direct way to control the agent's behavior by limiting its action space, reward shaping influences the learning process indirectly by modifying the reward signal. Each method has its strengths and weaknesses; action masking may provide faster convergence but at the potential cost of flexibility, whereas reward shaping allows for more nuanced guidance but can be more challenging to design effectively. Future research could focus on hybrid approaches that combine action masking with other RL enhancements, such as exploration rewards or traditional reward shaping techniques. Such a dual approach could leverage the strengths of both methods, promoting robust learning by both restricting irrelevant actions and encouraging desirable behaviors through carefully designed reward signals. This could help overcome some of the limitations associated with each method when applied in isolation, ultimately leading to more versatile and effective RL systems.

\newpage

\noindent
\textbf{Declaration of interests:} The authors declare that they have no known competing financial interests or personal relationships that could have appeared to influence the work reported in this paper.

\bibliographystyle{spmpsci_nat}

\bibliography{references}

\newpage
\section*{Appendix A: Details on Implementation and Hyperparameters}
\label{sec:appendix_ppo}

\renewcommand{\thetable}{A.\arabic{table}} % Prefix tables with "A."

\setcounter{table}{0}

\Cref{tbl:hyperparameter} presents the hyperparameter settings of PPO for all studies. We implement the PPO algorithm from the RL framework ``Stable Baselines 3'' version 1.5.0 \citep{raffin2021stable}. All parameters are set to their default values.

\begin{table}[H]
\caption{Hyperparameters of proximal policy optimization. \label{tbl:hyperparameter}}
\centering
{
\begin{tabular}{L{8cm}L{1.5cm}}
        \toprule
        \textbf{Hyperparameter}    & \textbf{Value}   \\ 
        \midrule
		Size of hidden layer & 64  \\ 
		Number of hidden layers  &  2   \\
		%Number of timesteps  & 10,000,000 \\
        Horizon & 2048 \\ 
        Clipping parameter ($\varepsilon$) & 0.20 \\
        State-value estimate coefficient ($c_1$) & 0.50 \\
        Entropy coefficient ($c_2$) &  0.00 \\ 
        Number of epochs &  10 \\ 
        Adam stepsize   &  0.0003\\ 
        Minibatch size &  64   \\ 
        Discount factor ($\gamma$) &  0.99  \\ 
        Generalized advantage estimate (GAE) parameter ($\lambda$)  &  0.95   \\ 
        \bottomrule
\end{tabular}
}
\end{table}

\end{document}